\documentclass[runningheads]{llncs}
\usepackage{graphicx}
\usepackage[misc]{ifsym}
\usepackage{todonotes}
\usepackage{blindtext,letltxmacro,xcolor,xparse}
\usepackage{lipsum}
\usepackage{siunitx}

\usepackage{caption}
\usepackage{subcaption}

\LetLtxMacro{\blindtextblindtext}{\blindtext}
\LetLtxMacro{\blindtextBlindtext}{\Blindtext}

\RenewDocumentCommand{\blindtext}{O{\value{blindtext}}}{%
  \begingroup\color{gray}\blindtextblindtext[#1]\endgroup
}
\RenewDocumentCommand{\Blindtext}{O{\value{blindtext}}O{\value{Blindtext}}}{%
  \begingroup\color{gray}\blindtextBlindtext[#1][#2]\endgroup
}

\begin{document}
\title{Towards Writing Style Adaptation in Handwriting Recognition}
%
%

\author{Jan Kohút (\Letter) \orcidID{0000-0003-0774-8903},
Michal Hradiš\orcidID{0000-0002-6364-129X},
\\and 
Martin Kišš\orcidID{0000-0001-6853-0508}}

\authorrunning{J. Kohút et al.}
%
\institute{Faculty of Information Technology, Brno University of Technology, Brno, Czech~Republic \\
\email{\{ikohut,ihradis,ikiss\}@fit.vutbr.cz}}
\maketitle              
\begin{abstract}
One of the challenges of handwriting recognition is to transcribe a large number of vastly different writing styles.
State-of-the-art approaches do not explicitly use information about the writer's style, which may be limiting overall accuracy due to various ambiguities. We explore models with writer-dependent parameters which take the writer's identity as an additional input.
The proposed models can be trained on datasets with partitions likely written by a single author (e.g. single letter, diary, or chronicle).
We propose a Writer Style Block (WSB), an adaptive instance normalization layer conditioned on learned embeddings of the partitions.
We experimented with various placements and settings of WSB and contrastively pre-trained embeddings.
We show that our approach outperforms a baseline with no WSB in a writer-dependent scenario and that it is possible to estimate embeddings for new writers.
However, domain adaptation using simple fine-tuning in a writer-independent setting provides superior accuracy at a similar computational cost.
The proposed approach should be further investigated in terms of training stability and embedding regularization to overcome such a baseline.
\end{abstract}

\begin{figure}[h]
    \centering
    \includegraphics[width=\linewidth, trim=15mm 245mm 60mm 37mm, clip]{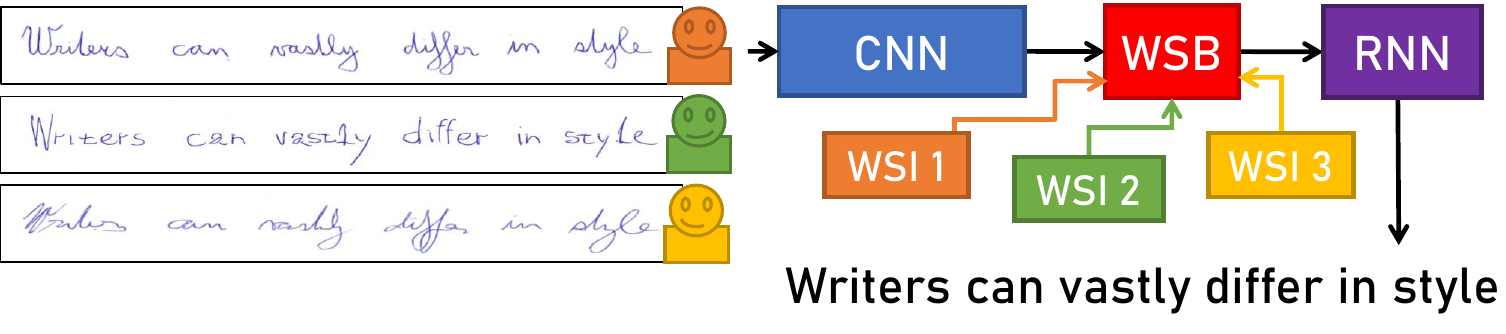}
    \caption{Our proposed Writer Style Block (WSB) learns to utilize various writer styles based on writer-style identifiers (WSI).}
    \label{fig:introduction:WS-Net}
\end{figure}

\section{Introduction}\label{sec:introduction}

Handwritten text of multiple writers can vastly differ in style, for example, the degree of slant, the way letters are joined (cursive or block letters), spacing between letters and words, similarity to printed text, the width of the stroke, etc.
In fact, some characters may not be recognizable without the knowledge of the writer's identity.
To achieve sufficient text recognition accuracy, the state-of-the-art neural networks~\cite{E2ENNCTC2015,AREMDLSTMNECESSARY2017,GatedConv2017,IMPROVINGCNNBLSTMCTC2018,Seq2Seq2019,PayAttentionTransformer2020,BiDecodingTransformer2021} must adapt to a large number of writer styles.
While transcribing, these architectures have to rely only on the image of a text line, which may not provide sufficient context.
The improper adaptation may lead to wrong interpretations of ambiguities, which naturally arise among multiple writers.

Figure~\ref{fig:introduction:WS-Net} illustrates WS-Net, which is a standard CTC-based~\cite{E2ENNCTC2015,AREMDLSTMNECESSARY2017,GatedConv2017,IMPROVINGCNNBLSTMCTC2018} architecture enhanced by our proposed Writer Style Block (WSB).
Apart from the text line image, WS-Net takes an additional input in the form of a Writer Style Identifier (WSI).
WSI serves as an index into a WSB writer-style embedding table, where each writer is represented by a single embedding.
WSB is an adaptive instance normalization~\cite{AdaIN2017} conditioned on writer-style embeddings.
The adaptive instance normalization can modulate how the network processes information and which features become important and as such it provides WS-Net the ability to adapt to a vast amount of different writers.

The specific contributions of this paper are as follows: 
(1) Writer Style Block (WSB), an adaptive instance normalization conditioned by writer-style embeddings, which can enhance any standard text handwritten recognition network with the ability to explicitly utilize writing styles in the training dataset; 
(2) extensive evaluation of WSB for various embedding dimensions 16--256, both in standard and pre-trained mode; 
(3) evaluation of WSB in writer-dependent and writer-independent scenarios.

\section{Related Work}\label{sec:related_work}




In the state-of-the-art approaches to handwritten text recognition, the writer's identity information is usually not utilized.
However, providing the recognition model with such information might be useful as the model can then better handle different writing styles, writer-specific patterns, etc.
Closely related methods to the model adaptation, such as transfer learning, fine-tuning, and other similar techniques, are studied in several works~\cite{FinetuningBaseline2023,Improving2019,OCR4AllHWRMedieval2022,OCR4AllPrinted2021,Seq2SeqAdaptation2019,TextAdaptation2019}, but the main disadvantage of these approaches is that they produce a unique model when adapted to a specific domain.
Our proposed Writer Style Block (WSB) enhances a text recognition network to explicitly use writer-style information, so it can handle multiple writers simultaneously.

An approach similar to ours was proposed by Wang et al.~\cite{WANG202242}. 
It consists of two neural networks -- writer-style extractor network and text recognition network, where the first network is trained to classify the writers' identities and the second one is trained to recognize the text content, each network accept text line image as input.
Local writer-style representations, obtained as output features of the last recurrent layer of the writer-style extractor network, are aggregated by a mean pooling layer into a global writer-style representation.
The writer's identity is utilized by aggregating the global writer-style representation and the local style representations by linear layers and connection operation into a single vector which is then aggregated into features of the text recognition network. 
When compared to our method, we do not extract the writer-style representation in a feed-forward manner, but instead, we learn a fixed number of representative embeddings on the training dataset, which allows the network to utilize all available writer-style information, not only the writer-style information of currently processed text line image.

Bhunia et al.~\cite{MetaHTR2021} proposed writer-style adaptation as a meta-learning task.
The goal is to train a general model which can be effectively adapted to a new writer, the adaptation should be fast and only a few labeled samples from the target domain should be needed.
They used a Seq2Seq model expanded of special meta parameters in the form of a gamma layer and dedicated learning rates for each layer.
During the adaptation process, the gamma layer should allow the model to focus more on problematic/unknown characters and ignore the already well-learned ones, while the dedicated learning rates should provide the model with the ability to prefer/ignore the adaptation of certain layers.
The meta parameters together with the general model parameters are optimized with a meta-learning process consisting of two phases, inner and outer.
The inner phase fine-tunes dedicated writer models (always initialized from the general one) on writer-specific data.
The outer phase evaluates the dedicated writer models on respective writer-held-out support sets and updates the meta parameters together with the general model's parameters with all the dedicated writer models gradients.
The inner and outer phases are repeated.
Instead of training a general model which can be effectively adapted to new writers, resulting in a dedicated model for each, our goal is to train a single model with writer-dedicated parameters, where the adaptation to a new writer is done by optimizing a new set of writer-dedicated parameters.


More extensive model adaptation research can be found in the speech recognition area (ASR)~\cite{SpeechAdaptationOverview2020}.
Structure transform approaches represent the most relevant domain adaptation methods to our work, the general idea is to build an architecture with a small set of speaker-dependent (SD) parameters for each speaker while keeping most of the parameters shared -- the speaker-independent (SI) parameters.
While training, both the SD and the SI parameters are updated, but during the adaptation to a new writer, only the SD parameters are optimized.
In previous works, the SD parameters include the input layer (linear input network, LIN~\cite{LIN1995}), the hidden layer (linear hidden network, LHN~\cite{LHN2007}), and the output layer (linear output network, LON~\cite{LON2010}).
Such adaptation has several drawbacks, mainly the large number of adapted parameters which results in a slow adaptation process and model overfitting if strong regularization is not used.
Also, as the speaker information is typically discarded in the latter layers of the network~\cite{SI2012}, in more recent approaches the SD parameters are located more toward the beginning of the network.
Many approaches aim to speed up the slow adaptation and suppress the overfitting problems mentioned above by reducing the number of adapted SD parameters.
Parametrization of activation functions with SD parameters was proposed by Zhang et al.~\cite{ParametrisedActivation2015}.
In other works, the SD parameters are represented by scales and/or offsets in various layers.
Namely, in Learning Hidden Unit Contributions (LHUC)~\cite{LHUC2016}, every kernel is followed by an SD scale parameter.
Another option is to use the scales and offsets in batch normalization layers as the SD parameters~\cite{BNOfflineAdaptation2017,BNOnlineAdaptation2019}.
Approaches by Zhao et al.~\cite{LRPD2016,eLRPD2017} and Samarakoon et al.~\cite{FHL2016} propose to factorize weight matrices of SD linear layers as most of the information is stored in diagonals.
Utilization of such decomposed matrices results in fewer SD parameters.

Approaches proposed by Cui et al.~\cite{EmbeddingBased2017} and Delcroix et al.~\cite{ContextAdaptive2018} use an auxiliary SI network that generates SD parameters based on a small SD input (e.g. i-vector, learned speaker embedding, or similar features).
In the first approach, the auxiliary network generates SD scales and offsets for hidden activations.
The latter approach proposes to train a recognition network with several branches under the assumption that the different branches learn to specialize in different types of speakers.
The auxiliary network is then used to generate weights for aggregation of the outputs produced by the individual branches.
Similarly to Wang et al.~\cite{WANG202242}, some of the existing methods~\cite{DynamicLayerNormalization2017,SpeakerOffset2019,LHUCOnline2019} 
 do not explicitly utilize the speaker's identity, instead, they extract global style features from the entire utterance, that is being currently recognized, and incorporate these features into the recognition process of its parts.

Our WS-Net architecture utilizes embeddings as writer-dependent (WD) parameters, while the rest of the parameters are shared among all writers (WI).
The writer-dependent embeddings are part of the Writer Style Block (WSB), which is an adaptive instance normalization connected to the writer-independent part of the WS-Net.
Because the adaptive instance normalization is conditioned on the writer-dependent embeddings, the information about a writer is utilized by the writer-independent part of the network.
WS-Net is inspired by style transfer approaches and style-dependent Generative Adversarial Networks, which use an adaptive instance normalization (AdaIN) to broadcast information about the desired output style across a whole image~\cite{CIN2016,AdaIN2017,AuxStyle2017,StyleGAN2018}.
Instead of using the AdaIN 
 layers to broadcast the information about the style, Murase et al.~\cite{AdaptiveAutoencoder2020} used it to broadcast the information about the content while optimizing an autoencoder for writer verification.
 In this way, the autoencoder can focus to extract only the writer-style information which is important for successful writer verification.
In our previous work~\cite{TS-Net2021} we introduced TS-Net, where the only difference to WS-Net is the location of the AdaIN layer.

\section{Writer Style Block}\label{sec:writer_style_block}

\begin{figure}[t]
    \centering
    \includegraphics[width=\linewidth, trim=95mm 85mm 20mm 0mm, clip]{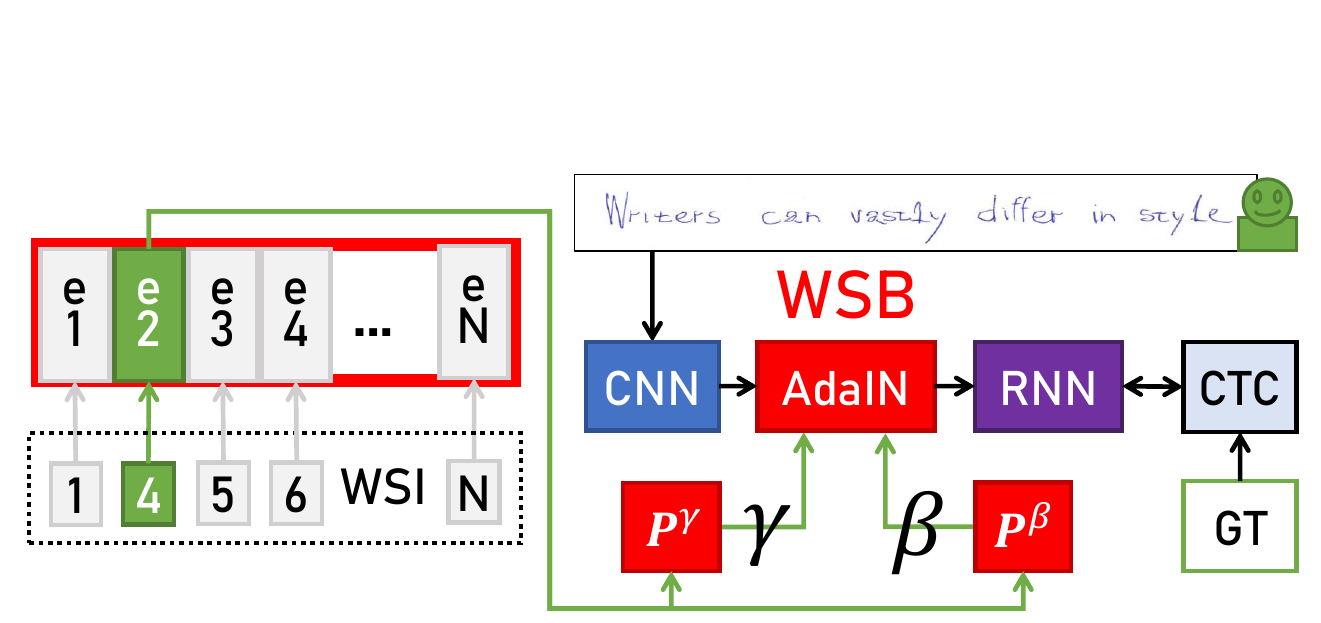}
    \caption{Our proposed neural network (WS-Net) consists of a convolutional part (CNN), a recurrent part (LSTM), and Writer Style Block (WSB). 
    }
    \label{fig:wsb:multi_writer_styles_model_wsb}
\end{figure}

We propose Writer Style Block (WSB), which allows WS-Net, described in detail in Section~\ref{sec:WS-Net}, to learn dedicated writer parameters in the form of writer-style embeddings.
Figure~\ref{fig:wsb:multi_writer_styles_model_wsb}
 shows WSB as part of the WS-Net.
The WSB is an adaptive instance normalization layer AdaIN~\cite{AdaIN2017}:

\begin{equation}
\mathrm{AdaIN}(\mathbf{X}_c, \gamma_c, \beta_c) = \gamma_c\bigg(\frac{\mathbf{X}_{c} - \mu(\mathbf{X}_{c})}{\sigma(\mathbf{X}_{c}) + \epsilon}\bigg) + \beta_c,
\end{equation}

where $\mathbf{X}$, $\mathbf{\gamma}$, $\mathbf{\beta}$, $\mu$, $\sigma$, $\epsilon$ stand for input, scales, offsets, mean, standard deviation and a small positive constant, $c$ specifies the channel dimension. 
The adaptive scales $\mathbf{\gamma}$ and offsets $\mathbf{\beta}$ are given by two affine projections $P^{\gamma}$ and $P^{\beta}$:

\begin{equation}
     \gamma = P^{\gamma}(\mathbf{e}) = \mathbf{W}^\gamma \mathbf{e} + \mathbf{{b}^\gamma}, \beta = P^{\beta}(\mathbf{e}) = \mathbf{W^\beta} \mathbf{e} 
     + \mathbf{b^\beta}, 
\end{equation}

where $\mathbf{W}^\gamma$, $\mathbf{W}^\beta$, $\mathbf{{b}^\gamma}$, $\mathbf{{b}^\beta}$ are projection matrices and biases and $\mathbf{e}$ is a writer-style embedding specified by the corresponding Writer Style Identifier (WSI).
Therefore, in our architecture, each writer has dedicated parameters in the form of a writer-style embedding $\mathbf{e}$, while all the other parameters are shared.
While training, a writer-style embedding $\mathbf{e}$ is updated only on the respective writer training data, all the other parameters of WS-Net are updated on all writers.

\subsubsection{Initialization.}
We initialize WSB similarly to the standard instance normalization.
Each writer-style embedding $\mathbf{e}$ is initialized from the standard normal distribution $\mathcal{N}(0,1)$. 
The projection matrices $\mathbf{W}^\gamma$, $\mathbf{W}^\beta$ are initialized from an uniform distribution $\mathcal{U}(-\sqrt{\mathrm{ED}}\times\tau, \sqrt{\mathrm{ED}}\times\tau)$, where ED stands for embedding dimension and $\tau$ is a small positive constant. 
Conditioning the initialization on the square root of the ED results in the same standard deviation of the scales $\mathbf{\gamma}$ and the offsets $\mathbf{\beta}$ across all ED.
The magnitude of this standard deviation can be manipulated with $\tau$.
In conducted auxiliary experiments we found that a reasonable value of the standard deviation is 0.1 and it is achieved by setting $\tau$ to 0.174.
The $\mathbf{{b}^\gamma}$ and $\mathbf{{b}^\beta}$ are set to ones and zeros, respectively.
For all ED, this way of initialization ensures, that both the scales $\mathbf{\gamma}$ and the offsets $\mathbf{\beta}$ have the same small standard deviation, while the scales are centered around 1 and the offsets around 0.
Instead of initializing the embeddings from the standard normal distribution, pre-trained embeddings can be used (see Section~\ref{sec:pretrained_embeddings}).

\section{CzechHWR Dataset}\label{sec:dataset}

Our dataset CzechHWR consists of triplets: a text line image, WSI, and the ground-truth transcription of the text line image.
It was created from three main sources: our text recognition web application PERO OCR\footnote{https://pero-ocr.fit.vutbr.cz}, a collection of Czech letters~\cite{hladka_2013}, and Czech chronicles.
So far, users of the web application have uploaded and annotated documents containing about 295k handwritten text lines.
Most of the documents were written in Czech modern cursive script, although, a marginal amount of documents in different scripts such as Gothic or German Kurrent is also present.
They are mainly composed of military diaries, chronicles, letters, and notes.
Based on document or page level manual inspection of the annotated documents, we assigned approximately 2.6k WSI.
Czech letters is a collection of 2k letters, where the number of text lines is 87k.
Most of them were written in Czech modern cursive, while a minimum amount was typeset.
As it can be reasonably assumed that a handwritten letter has only one author, we assigned a distinct WSI to each letter.
We manually annotated approximately 2 pages of 277 distinct Czech chronicles, resulting in 553 annotated pages with 24k text lines. 
As each page was chosen from a visually different place in the chronicle, we assigned a different WSI to each page.

\begin{figure}[t]
    \centering
    \includegraphics[width=\linewidth, trim=15mm 95mm 80mm 8mm, clip]{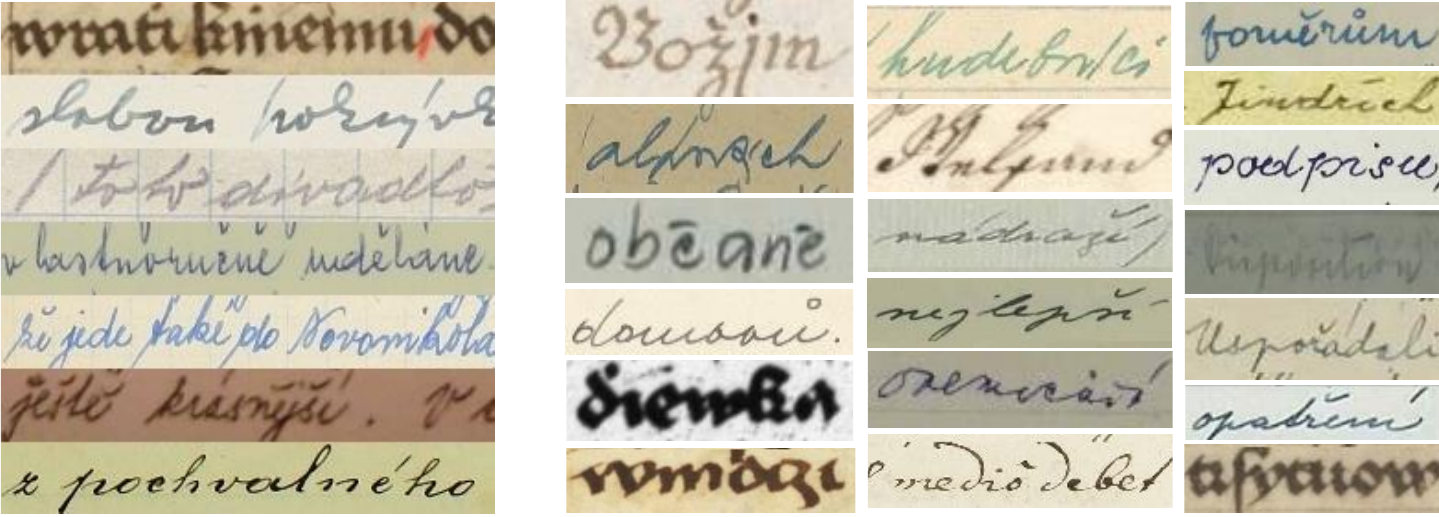}
    \caption{Left, samples from the CzechHWR dataset. Right, representative words of 19 writers from Handwriting Adaptation Dataset.}
    \label{fig:experiments:dataset:example}
\end{figure}
The final CzechHWR dataset contains 406k annotated text line images with 5.1k WSI.
The level of penmanship/readability differs, ranging from scribbles to calligraphy, although due to the origin of the data, the tendency is towards fairly readable texts (see the left side of Figure~\ref{fig:experiments:dataset:example}).
There are two issues resulting from the WSI assigning process.
First, there is no assurance that multiple WSI does not identify the same writer because it is not possible for the annotator to keep track of thousands of writing styles.
Second, as we assigned the WSI on the document or page level, there is a possibility, that some text lines of the document/page belonged to distinct writers. 
The first issue should not present a problem for our system, as it will try to learn the same writer-style embedding $\mathbf{e}$ multiple times.
The second issue should present only a slight regularization to the optimization process, as the vast majority of text lines were assigned correctly. 
The exact number of distinct writers is unknown, but a reasonable lower estimate is 4.5k.
To avoid overcomplicated statements we often use ``writer lines'' to refer to data samples (triplets) that share the same WSI.

We divided the CzechHWR dataset for training and testing in the following way.
We randomly draw 5k lines for testing (TST).
The remaining lines were divided into seven clusters: 1, 22, 55, 110, 225, 550, and 1100, according to their writers (WSI). 
The cluster numbers specify the minimum number of lines for each writer to belong to that cluster.
A writer belongs only to the cluster with the highest possible number, the resulting clusters are disjoint. 
Out of clusters 22, 55, 110, 225, 550, and 1100, for each writer (WSI), we respectively took 2, 5, 10, 25, 50, and 100 lines for testing clusters $\mathrm{TST_C}$.  
The remaining lines formed the training clusters $\mathrm{TRN_C}$.
Out of convenience, we refer to both testing and training clusters with the minimum number of lines belonging to each writer in the training clusters: 1, 20, 50, 100, 200, 500, and 1000, e.g. $\mathrm{TST_{20}}$ or $\mathrm{TRN_{500}}$.
For example, a CER measured on $\mathrm{TST_{50}}$ set shows an error for writers that have at least 50 lines in the training dataset. 
The number of lines and WSI for each cluster is shown in Table~\ref{tab:experiments:dataset:dataset},
ALL stands for all clusters.
\begin{table}[t]
\caption{The number of lines and writers (WSI) for each cluster in the CzechHWR dataset. 
        For a detailed description, see the text. }\label{tab:experiments:dataset:dataset}
\centering
{
\begin{tabular}{r | p{10mm} p{10mm} p{10mm} p{10mm} p{10mm} p{10mm} p{10mm} | l}
& 1 & 20 & 50 & 100 & 200 & 500 & 1000 & ALL \\
\hline\hline 
TRN & 13k & 79k & 82k & 43k & 24k & 16k & 122k & 379k \\
TST & 169 & 1k & 1.1k & 566 & 287 & 198 & 1.7k & 5k \\ 
$\mathrm{TST_W}$ & 0 & 4.5k & 6.2k & 3.2k & 2k & 1.1k & 5.4k & 22.4k \\
\hline
WSI & 1.1k & 2.3k & 1.2k & 322 & 79 & 21 & 54 & 5.1k \\
\end{tabular}}
\end{table}
For training, we merged all training clusters into a single training dataset (TRN). 
TST allows us to inspect the average CER because the distribution of WSI is the same as in TRN.
$\mathrm{TST_C}$ and $\mathrm{TRN_C}$ clusters allow us to measure how the number of data samples per WSI affects the WSB performance.

For experiments with writer-independent scenario we used our Handwriting Adaptation Dataset\footnote{https://pero.fit.vutbr.cz/handwriting\_adaptation\_dataset}~(HAD)~\cite{FinetuningBaseline2023} which consists of 19 writers, right side of Figure~\ref{fig:experiments:dataset:example} shows sample words for each writer.
In comparison with CzechHWR dataset, HAD contains both similar and vastly different scripts.

\section{Writer Style Network}\label{sec:WS-Net}

In this section, we describe the proposed WS-Net and its training procedure.
We first introduce the baseline network architecture and later we specify the changes that lead to the proposed WS-Net.

The baseline architecture is similar to text recognition state-of-the-art architectures trained with CTC loss function~\cite{E2ENNCTC2015,AREMDLSTMNECESSARY2017,GatedConv2017,IMPROVINGCNNBLSTMCTC2018}.
It consists of a convolutional part (CNN) and a recurrent part (RNN).
The CNN part is a sequence of 4 convolutional blocks, where each has 2 convolutional layers with numbers of output channels set to 64, 128, 256, and 512, respectively.
All convolutional blocks except the last one are followed by a max pooling layer, while the input width subsampling factor of the CNN is 4.
The RNN part processes three scaled versions of the WSB output with three branches, the scaling factors are 1, 0.5, and 0.25 and each branch has two LSTM layers.
The outputs are upsampled back to the original resolution, summed, and processed with a final LSTM layer.
Each LSTM layer is bidirectional and has a hidden feature size of 256 for each dimension.
The output of the RNN block is processed by a final 1D convolutional layer.
The baseline architecture has 5 instance normalization layers, the first four are in the convolutional part after each convolutional block (CNN), and the last is after the recurrent block (RNN).
A more detailed description together with a detailed diagram can be found in our previous work~\cite{TS-Net2021}.

WS-Net (see Figure~\ref{fig:wsb:multi_writer_styles_model_wsb}) is the baseline architecture enhanced with WSB, which replaces one or multiple standard instance normalization layers with adaptive instance normalization layers (AdaIN) conditioned on writer embeddings.
Each AdaIN has its own projection matrices $\mathbf{W}^\gamma$, $\mathbf{W}^\beta$, and biases $\mathbf{{b}^\gamma}$, $\mathbf{{b}^\beta}$, but the writer-style embeddings $\mathbf{e}$ are shared.
We experimented with two variants of WS-Net: Single AdaIN and All AdaIN.
Single AdaIN is WS-Net, where the adaptive normalization layer (AdaIN) is placed after the convolutional block (CNN), and all the rest are standard instance normalization layers.
All AdaIN is WS-Net, where every normalization layer is AdaIN.

\subsubsection{Motivation behind AdaIN placements.}
The motivation behind the AdaIN layer placement in Single AdaIN architecture is based on auxiliary experiments with All AdaIN architecture trained in multiple embedding dimension (ED) settings.
By fixing scales and offsets for various AdaIN layers of a trained All AdaIN system, we simulated all possible settings of AdaIN layers.
We fixed the respective AdaIN by conditioning the scales and offsets on the mean writer embedding.
Out of all placements, we noticed a significantly poorer performance for every setting where the AdaIN layer after the CNN block was fixed, which suggests that the All AdaIN system benefited most from this adaptive normalization layer.
Furthermore, in speech recognition, A Mohamed et al.~\cite{SI2012} found that information about speakers generally vanishes toward the end of the network, which suggests that placing the AdaIN layer near the end of the network might have no effect.
On the other hand, placing the AdaIN near the begging of the network might have no effect too as the activation in the first layers usually bears only low-level information.
Finding the best possible setting for AdaIN layers would mean training 31 different settings from scratch, for each embedding size ED.
Additionally, we experimented with a setup where the AdaIN layer was behind the CNN and the RNN block, which turned up to be unstable and therefore we do not discuss the respective results in more detail.

\subsubsection{Training.} WS-Net is trained jointly with Adam optimizer~\cite{Adam2015}, and the CTC loss function. 
The training data consists of triplets: a text line image, WSI, and the ground-truth transcription.
Because the training samples are drawn randomly, embeddings that have more data samples assigned to them are updated more frequently.
We mitigate this by normalizing embedding batch gradients by the frequency of their WSI in the batch. 
We trained Single AdaIN and All AdaIN, with the embeddings initialized from the standard normal distribution (normal embeddings), for 500k iterations up until convergence.
We used polynomial warmup of a third order to gradually increase the learning rate from 0 to \num{3e-4} in the first 10k iterations.
At iterations 200k and 400k, we used the warmup again, but the learning rate maximums were \num{7e-5} and \num{1.75e-5}.
The batch size was set to 32. 
The baseline was trained with the same strategy. 

Additionally, we trained Single AdaIN and All AdaIN, with the pre-trained embeddings (described in Section~\ref{sec:pretrained_embeddings}), for 675k iterations in three consecutive steps.
For each step, we trained the model up until convergence.
First, we optimized all the parameters except the embeddings for 400k iterations.
We used the warmup strategy in iterations 0, 200k, and 300k with the learning rate maximums being \num{3e-4}, \num{1.5e-4}, and \num{7e-5}.
Second, for the next 100k iterations, we optimized just the embeddings.
We used the warmup strategy in iterations 400k and 450k and kept the learning rate the same.
Third, for the final 175k iterations, we finetuned all the parameters.
We used the warmup strategy at iterations 500k, 550k, 600k, and 650k with the learning rate maximums being \num{7e-5}, \num{3e-5}, and \num{1.5e-5}. 
The model was significantly more accurate after each step.

We used data augmentation including color changes, noise, blur, and various geometric transformations to simulate different backgrounds, slants, spacing between characters, etc.
Additionally, we mask the text line images with a random number of noise patches.
The height of a noise patch is the same as the height of the text line image, the width is chosen randomly up to the height of the text line image forming at most a square patch.
In this setting, the noise patch usually masks a maximum of two letters.
The intuition behind masking is to strengthen the language modeling capability of the system.
A more detailed description of the used augmentation together with examples can be found in our work~\cite{FinetuningBaseline2023}, where the respective augmentation is B1C1G1M1.

\section{Pre-training Writer-Style Embeddings}\label{sec:pretrained_embeddings}

Instead of using embeddings initialized from the standard normal distribution, we also use pre-trained ones.
We implemented a contrastive learning approach, where the encoder is a stack of four convolutional layers with output channel dimensions 32, 64, 128, and 512 respectively, followed by three multi-head attention blocks with 4 heads and 512 channels, average pooling over the width dimension, and L2 normalization.
The encoder generates an embedding $\mathbf{q}$ for each text line image input.
We used the normalized
temperature-scaled cross entropy (NT-Xent) loss function~\cite{NTXENT2020}:

\begin{equation}
\mathcal{L}_{\mathbf{qp}} = -log\frac{\mathrm{exp}(\mathbf{q}\cdot\mathbf{p}/\tau)}{\sum_{j=0}^{N}\mathrm{exp}(\mathbf{q}\cdot\mathbf{n}_j/\tau) + \mathrm{exp}(\mathbf{q}\cdot\mathbf{p}/\tau)},
\label{eq:wsb:ntxent}
\end{equation}

where $\mathbf{q}$ forms a positive pair with the embedding $\mathbf{p}$, and  negative pairs with embeddings $\mathbf{n}_j$.
Embeddings of a positive pair are generated from image text lines belonging to the same writer, while embeddings of a negative pair are generated from image text lines belonging to distinct writers.
The final NT-Xent loss is given by $\mathcal{L} = \mu(\mathcal{L}_{\mathbf{qp}}), (\mathbf{q}, \mathbf{p}) \in \mathbf{P}$, where $\mathbf{P}$ is a set of all positive pairs.
As the final layer of the encoder is L2 normalization, the dot product will produce cosine similarity.
The temperature parameter $\tau$ affects the strictness of the NT-Xent loss function.
Values closer to 0 make it more strict, whereas higher values make it looser.
Higher strictness force the encoder to produce closer cosine similarities, which means closer embeddings. 
We set $\tau$ to 0.15.

As the encoder can only provide an embedding $\mathbf{q}$ for a text line image, we extract the final writer-style embedding $\mathbf{e}$ by aggregating output embeddings for 32 distinct and random text line images belonging to the respective writer.
The aggregation is done by choosing the embedding $\mathbf{q}_i$ which has the largest sum of above-average cosine similarities.
The cosine similarity of two embeddings $\mathbf{q}_i$ and $\mathbf{q}_j$ is above average if it is larger than the average cosine similarity between all 32 embeddings.
No data augmentation is used during the extracting process. 

\subsubsection{Training writer-style encoder.}
The encoder was trained with AdamW~\cite{AdamW2017} optimizer for 20k iterations, with a batch size of 180, and learning rate \num{2e-4}. 
We use the same dataset as for the standard training, the augmentations are similar, but stronger, without any geometry transformation and patch noise masking.
The NT-Xent loss is evaluated for all positive pairs in the batch. We checked the convergence by visual inspection of text line images that belonged to the k-nearest neighbors embeddings according to cosine similarity.

\section{Writer-dependent Scenario}
\label{sec:experiments}



\begin{figure*}[t]
     \centering
     \begin{subfigure}[b]{0.5\linewidth}
        \centering
        \includegraphics[width=\linewidth, trim=0mm 0mm 0mm 0mm, clip]{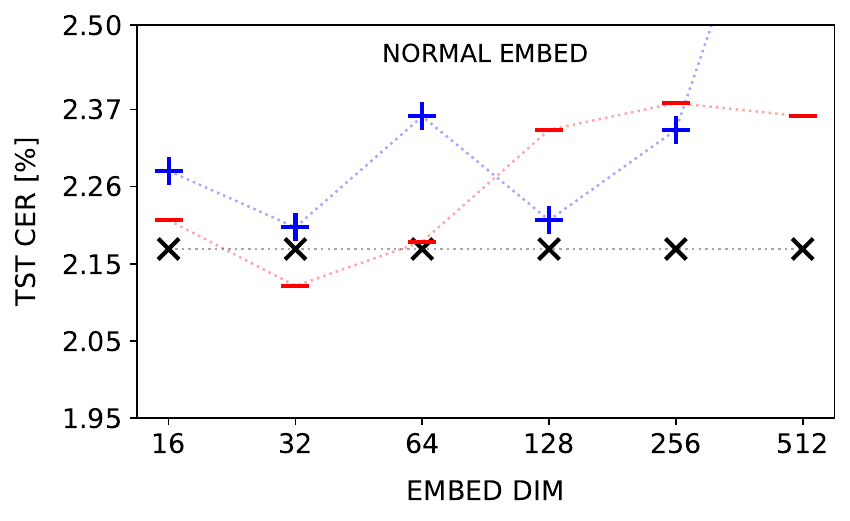}        \label{fig:experiments:baseline:ED_SS_E}
     \end{subfigure}%
     ~
     \begin{subfigure}[b]{0.5\linewidth}
         \centering
        \includegraphics[width=\linewidth, trim=0mm 0mm 0mm 0mm, clip]{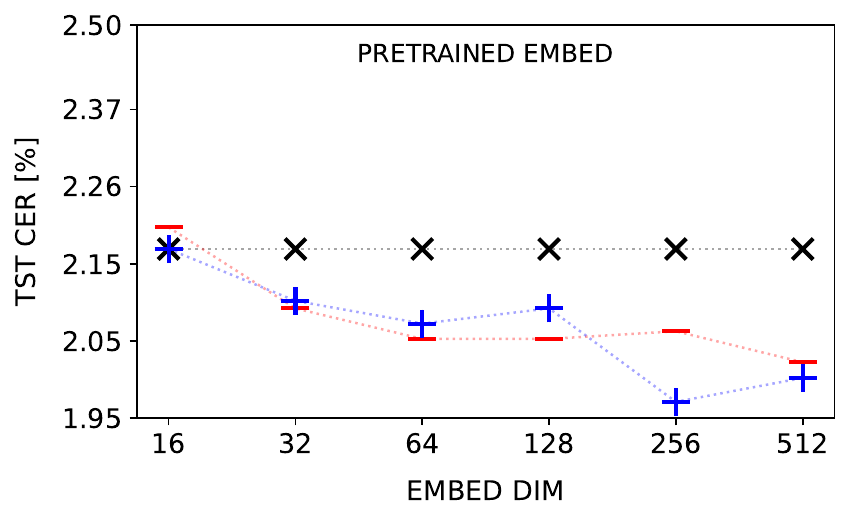}        \label{fig:experiments:baseline:ED_PP_E}
     \end{subfigure}

        \vspace{-5mm}
     
     \begin{subfigure}[b]{0.5\linewidth}
         \centering
        \includegraphics[width=\linewidth, trim=0mm 0mm 0mm 85mm, clip]{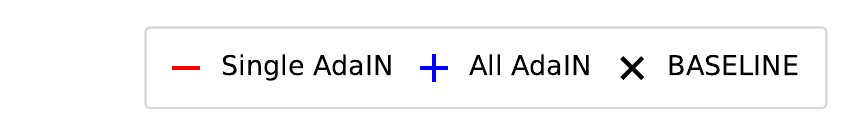}
        \label{fig:experiments:baseline:ED_legend}
     \end{subfigure}

             \vspace{-5mm}

        \caption{Character error rate (CER) for Single AdaIN, All AdaIN, and the baseline on the test set. The graphs show CER for different embedding dimensions (ED) and for different initialization: randomly initialized (left) and pre-trained (right).}
        \label{fig:experiments:baseline:ED}
\end{figure*}
In this section, we describe the experiments conducted with the WS-Net on the CzechHWR dataset.
Specifically, we compare Single AdaIN and All AdaIN variants of WS-Net to the baseline network.
We trained architectures in both normal and pre-trained embedding setups for embedding dimensions (ED): 16, 32, 64, 128, 256, and 512, and a separate writer-style encoder was trained for each ED.
Figure~\ref{fig:experiments:baseline:ED} shows test character error (CER) for Single AdaIN, All AdaIN, and the baseline on the testing set TST. 
The left graph shows architectures initialized with normal embeddings.
The Single AdaIN performed better for ED 16, 32, 64, and 512, and brought the best result for ED 32, All AdaIN was inconsistent across ED.
All settings, except Single AdaIN with ED 32, brought worse performance than the baseline.
The right graph shows architectures initialized with pre-trained embeddings.
All settings consistently outperformed the baseline, with the exception of ED 16, which brought similar CER.
Both AdaIN settings brought similar results, except for the ED 256, where All AdaIN had the lowest CER of all settings and decreased the test CER of the baseline by 9.22\% relatively.
All AdaIN is more stable across different ED, while initialized with pre-trained embeddings.
When initialized with pre-trained embeddings, Single AdaIN brought progressively better performance with increasing ED, whereas initialization with normal embeddings had the opposite tendency since the ED 32.
In further experiments, we do not show results for All AdaIN, as it did not bring any significant improvement over Single AdaIN.

     
     

     

     

\begin{figure}[t]
     \centering   
     \begin{subfigure}[b]{0.5\linewidth}
         \centering
        \includegraphics[width=\linewidth, trim=0mm 10mm 0mm 0mm, clip]{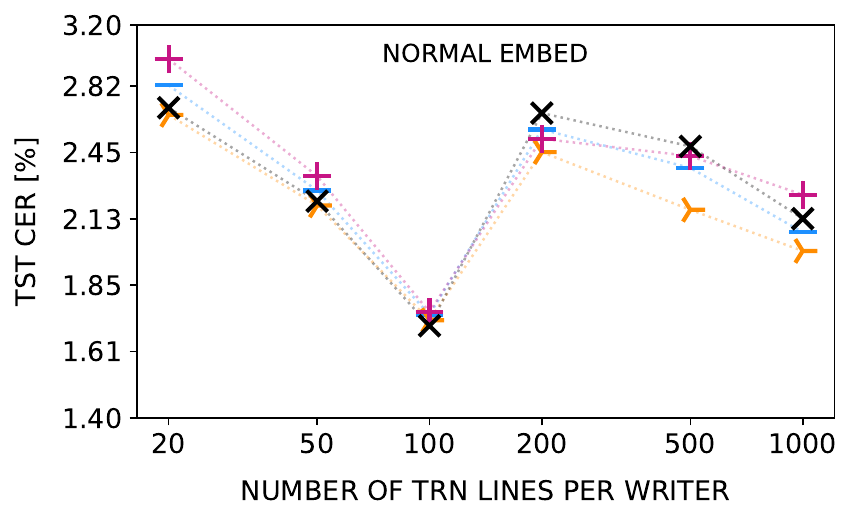}        \label{fig:experiments:baseline:TSTW_S_E_16_32_256}
     \end{subfigure}%
     ~
     \begin{subfigure}[b]{0.5\linewidth}
         \centering
        \includegraphics[width=\linewidth, trim=0mm 10mm 0mm 2mm, clip]{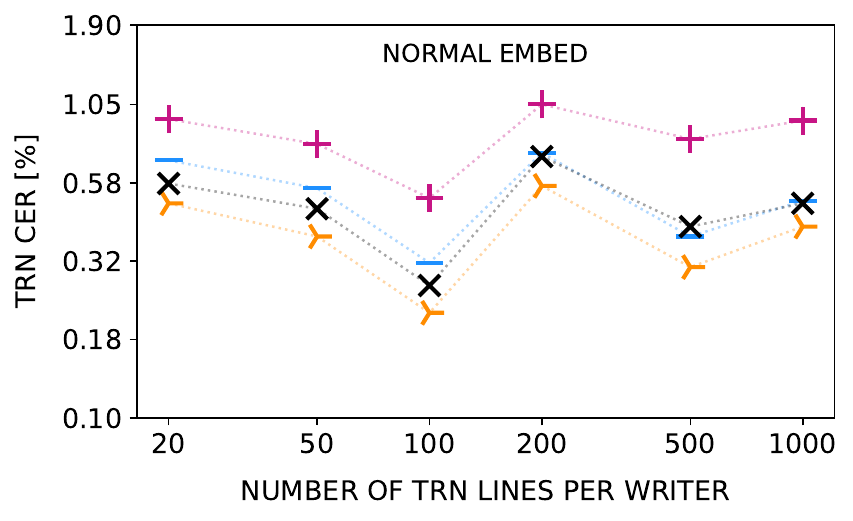}        \label{fig:experiments:baseline:TRN_S_E_16_32_256}
     \end{subfigure}
     
    \vspace{-3mm}
     
     \begin{subfigure}[b]{0.5\linewidth}
         \centering
        \includegraphics[width=\linewidth, trim=0mm 0mm 0mm 0mm, clip]{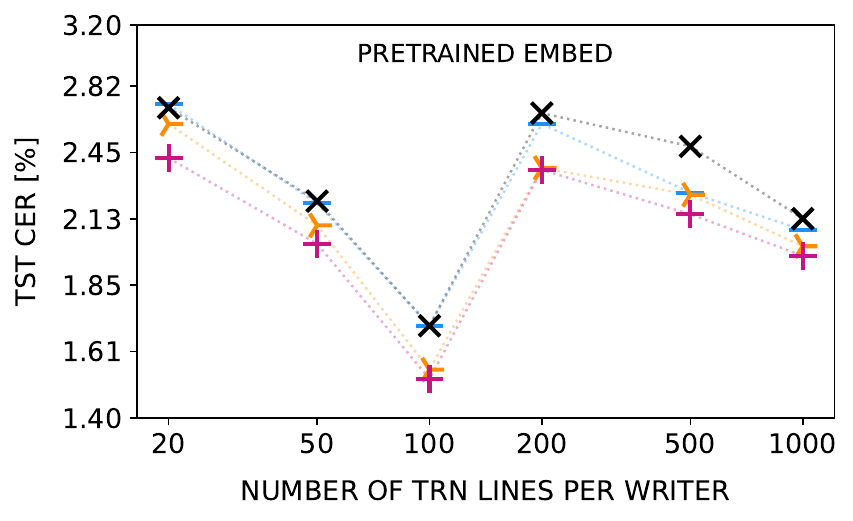}
        \label{fig:experiments:baseline:TSTW_P_E_16_32_256}
     \end{subfigure}%
     ~
     \begin{subfigure}[b]{0.5\linewidth}
         \centering
        \includegraphics[width=\linewidth, trim=0mm 0mm 0mm 0mm, clip]{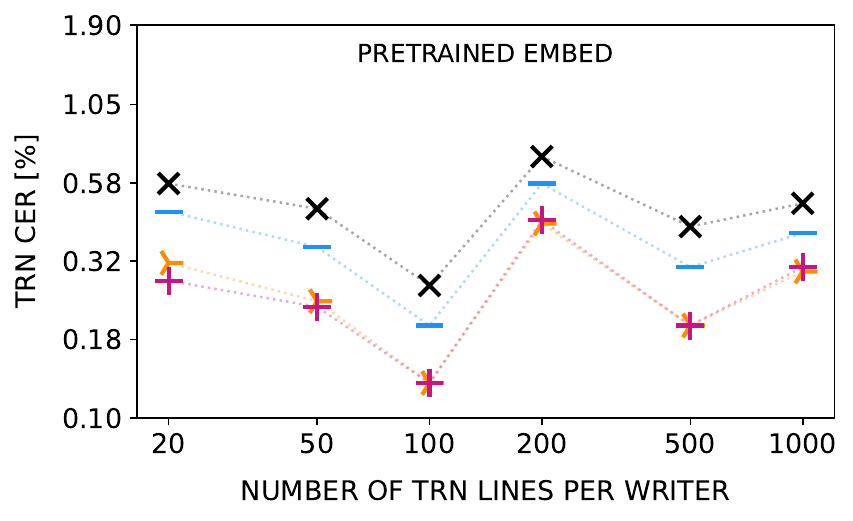}
        \label{fig:experiments:baseline:TRN_P_E_16_32_256}
     \end{subfigure}

    \vspace{-5mm}
     
     \begin{subfigure}[b]{\linewidth}
         \centering
        \includegraphics[width=0.5\linewidth, trim=0mm 0mm 0mm 85mm, clip]{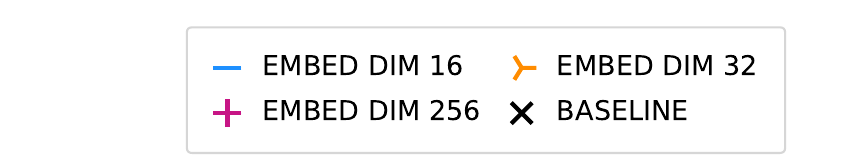}
        \label{fig:experiments:baseline:TSTW_E_16_32_256_legend}
     \end{subfigure}

         \vspace{-3mm}
     
        \caption{Character error rate (CER) measured on various testing and training clusters for Single AdaIN with randomly initialized embeddings (top) and pre-trained embeddings (bottom).}
        \label{fig:experiments:baseline:TSTW_E}
\end{figure}

Figure~\ref{fig:experiments:baseline:TSTW_E} shows Single AdaIN CER measured on various testing and training clusters (see Table~\ref{tab:experiments:dataset:dataset} and Section \ref{sec:dataset}) for ED 16, 32, and 256.
The top graphs show the CER for Single AdaIN initialized with normal embeddings.
We do not show the results for ED 64, 128, and 512, as the CER for ED 64 had the same tendency as ED 32, whereas ED 128 and 512 had the same tendency as ED 256.
For larger clusters 100, 200, 500, and 1000, the test CER was smaller or similar to the baseline, while for smaller clusters 20 and 50, it was worse or similar.
Only ED 32 and 64, outperformed the baseline for the larger clusters and evened out the baseline for smaller ones.
As there was no noticeable relative decrease in train CER for smaller clusters (20, 50), the respective writer-style embeddings probably overfitted in the wrong way and therefore brought poor generalization.
Although we trained ED 256 up until convergence, the train CER suggests that it was not trained properly, as it should have overfit more than the smaller ED and the baseline (the same applied for ED 128 and 512).


The bottom graphs in Figure~\ref{fig:experiments:baseline:TSTW_E} show the results for Single AdaIN initialized with pre-trained embeddings.
For all clusters, all ED except 16 had consistently smaller test CER than the baseline.
The general tendency among all clusters, both for the testing and the training clusters, is that a higher ED has always lower CER than a smaller ED.
Surprisingly, ED 16 brought significantly better performance for writers in cluster 500, this is probably due to the fact that the writer-style encoder was able to find the unique properties of their writing styles and encode them even to smaller embeddings.
Although all ED were fairly overfitted, they were able to generalize well among all the clusters.

The pre-trained embedding initialization variants generally outperformed the normal ones, while the largest differences were on smaller clusters.
If we compare the best ED out of each variant, ED 32 for normal embeddings to ED 256 for pre-trained embeddings, we can see that the latter was better only on the smaller clusters.
Single AdaIN trained from scratch (initialized with normal embeddings) does not guarantee in any way that the learned embeddings would represent the respective writer styles and therefore it is prone to overfit on writer irrelevant details, especially for smaller amount of writer lines.
On the other hand, Single AdaIN initialized with the pre-trained embeddings is forced to learn the proper utilization of writer styles in relation to the handwritten text recognition task, as the pre-trained embeddings are fixed for the first 400k training iterations.
As the writer-style encoder was directly trained to encode the writing style, it learned to extract robust and representative embeddings even for writers from smaller clusters.

For pre-trained embeddings and normal embeddings with ED 128 and 256, the t-SNE projections showed semantically meaningful clusters, where writers of similar scripts were grouped together.
For normal embeddings with ED 16, 32, and 64, the projections did not show visible clusters.

\section{Writer-independet Scenario}

For new writers, our architecture cannot be used in a simple feed-forward manner.
However, as all parameters except the writer-style embeddings are shared among the writers, we should be able to adapt to a new writer by finding a new representative embedding.
We experimented with two approaches.
The first selected the new embedding out of the existing ones according to CER on adaptation lines
As there are more than 5k existing embeddings, we clustered the existing embedding space with the k-mean algorithm into 50 clusters and evaluated only one random embedding from each cluster.
The second optimized a new embedding with 150 LBFGS iterations, the adaptation text line images were augmented in the same way as the training ones.
To inspect the quality of selection and optimization for different numbers of adaptation lines, we define writer adaptation runs.
A writer adaptation run consists of adapting the respective writer on 5 line clusters: 16, 32, 64, 128, and 256, where the numbers refer to the number of lines in them, a smaller cluster is always a subset of a larger one, and the lines of the largest cluster are drawn randomly from all available lines.
We run 23 adaptation runs for 19 new writers of HAD dataset and 3 fully-trained Single AdaIN architecture setups: normal embeddings with ED 32, pre-trained embeddings with ED 32, and pre-trained embeddings with ED 256, resulting in final $23\times19\times3$ runs for both the selection and optimization approach.
We always chose the best-performing embedding on the adaptation lines and inspected CER on test lines, there are 256 randomly drawn testing lines for each adaptation run.

The selection approach performed worse than the baseline even for higher amounts of adaptation lines.
Generally, for all Single AdaIN setups, the selection was more accurate for higher numbers of adaptation lines.
Surprisingly, the best-performing setup of Single AdaIN was normal embeddings ED 32, while for some writers and selections based on 256 adaptation lines, it performed similarly to the baseline.
We do not show detailed results, as the selection approach did not outperform the baseline and we did not notice any interesting properties apart from the already described.
For the optimization approach, we tried to optimize from the selected embeddings, but the mean of the existing ones turned out to be a better starting point.

\begin{figure*}[t]
    \centering
    \includegraphics[width=\textwidth, trim=0mm 0mm 0mm 0mm, clip]{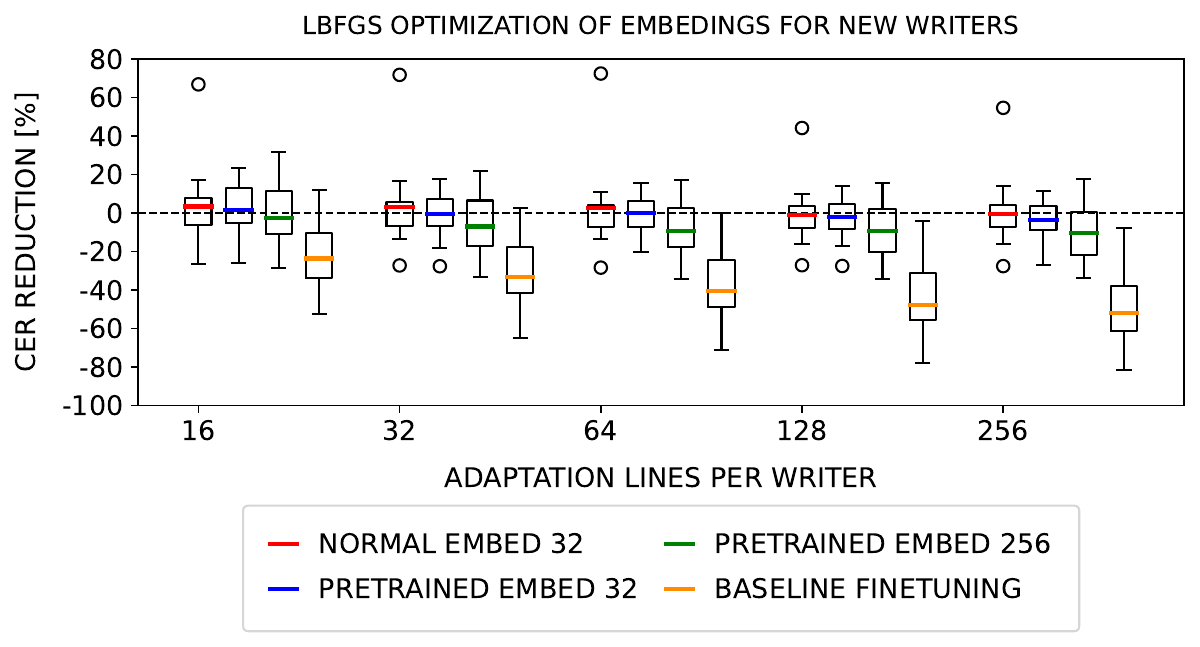}
    \caption{The performance of our optimization approach expressed as relative test CER reductions of the baseline compared to baseline finetuning.}
    \label{fig:experiments:LBGS:BM}
\end{figure*}
Figure~\ref{fig:experiments:LBGS:BM} shows the performance of our optimization approach expressed as relative test CER reductions of the baseline.
A boxplot represents the distribution of the 19 writers' CER reductions, and a writer's CER reduction is the mean of the writer's CER reductions measured on 23 runs.
More precisely, the CER reduction is given by:
\begin{equation}
\frac{A-B}{B}, 
\end{equation}
where A is the test CER of the adapted Single AdaIN, and B is the test CER of the baseline. 
Generally, for all Single AdaIN setups, the optimization was more accurate for higher numbers of adaptation lines.
By further inspecting the results in relation to Figure~\ref{fig:experiments:dataset:example}, we noticed that the performance varied across writers in relation to their scripts.
CER reductions were largest for scripts that were not sufficiently represented in the CzechHWR dataset, such as German Kurrent or Ghotic, while the performance worsen for the CzechHWR-like scripts.
On average, the pre-trained embeddings ED 256 setup provided the best performance, although it performed worse than the respective ED 32 setup for some writers.
The normal embeddings ED 32 setup brought the lowest CER reductions, while it vastly worsen the performance for some writers.

Based on these results, WS-Net is able to learn writer-style space that can significantly boost the transcription accuracy of new writers, however, finding representative embeddings is not straightforward.
We argue that this is to some extent caused by the sensitivity of WS-Net to precise writer-style embeddings.
We evaluated the sensitivity of WS-Net by randomly shuffling WSI in testing datasets, the test CER increased 2 times for the normal setting and 4 times for the pre-trained setting.
Therefore, simply selecting an existing embedding brought worse performance for all new writers, as WS-Net was too overfitted to existing writers and none of them had an extremely similar style to the new ones.
Optimization was difficult and unstable, as both the mean and the selected embeddings did not provide a good starting point.

Note that for Single AdaIN pre-trained ED 256 setup we evaluated WS-Net with embeddings provided by the writer-style encoder in an unsupervised manner, both without and with further optimization, but the results were comparable to supervised selection and optimization from the mean.
Furthermore, as the embeddings might have slightly changed in the last phase of the training, we tried to boost the performance of this approach by finetuning the writer-style encoder on the WS-Net writer-style embeddings using the L2 loss, but even this setup failed.

So far, the only reasonable solution for new writers is to find representative embeddings with optimization on larger numbers of adaptation lines, however
for such scenarios, a simple finetuning of the baseline brought significantly better test CER reductions.
Note that we estimated the optimal number of finetuning iterations with 4-fold cross-validation.
An extensive analysis of the finetuning approach can be found in our work~\cite{FinetuningBaseline2023}, where we showed that significant CER reductions can be obtained even with less than 16 adaptation lines. 
So far, WS-Net is only suitable for the writer-dependent scenario where it consistently outperformed the baseline, whereas it is not a suitable choice for writer-independent scenarios, especially if enough annotated data for new writers are available.

In our future work, we plan to redesign the WSB block, so it can be used in writer-independent scenarios without any adaptation lines.
The idea is based on an attention mechanism, which would take the hidden features of the processed text line image as queries, and embeddings of the writer-style space as keys and values. 
In speech recognition, similar ideas were proposed by Zhao et al.~\cite{SpeechMemory2020} and Fan et al~\cite{SpeakerAware2019}.

\section{Conclusion}

We showed that a standard CTC-based neural network enhanced by proposed
Writer Style Block (WSB) can utilize a vast number of writing styles.
While initializing WSB with embeddings pre-trained in an unsupervised contrastive manner, the enhanced architecture was able to consistently outperform the baseline version even for writers that were poorly represented in the training set.
On the other hand, training the WSB writer-style embeddings from scratch, led to worse performance on average.
Although we were unable to find an appropriate way to estimate representative embeddings for new writers, we confirmed their existence in the WSB writer-style embedding space by optimization on 256 adaptation lines.
This suggests that the WSB provides superior performance even for new writers.
In future work, we plan to learn the appropriate estimation in a supervised manner by extending the WSB with an attention mechanism, which would take the hidden features of a processed text line image as queries and the writer-style embeddings as keys and values.

\subsubsection{Acknowledgment.} This work has been supported by the Ministry of Culture Czech Republic in NAKI III project Machine learning for printed heritage digitisation (DH23P03OVV066).

%
%

\bibliographystyle{splncs04}
\bibliography{mybibliography}

\end{document}